\documentclass[10pt,twocolumn,letterpaper]{article}

\usepackage{cvpr}
\usepackage{times}
\usepackage{epsfig}
\usepackage{graphicx}
\usepackage{amsmath}
\usepackage{amssymb} 

\usepackage[linesnumbered,boxed]{algorithm2e}

\usepackage[dvipsnames,usenames]{color}
\usepackage[percent]{overpic}

\usepackage[pagebackref=true,breaklinks=true,letterpaper=true,colorlinks,bookmarks=false]{hyperref}

\cvprfinalcopy 

\def\cvprPaperID{1403} 

\ifcvprfinal\fi
\begin{document}





\title{ Egocentric Pose Recognition in Four Lines of Code}

\author{Gregory Rogez$^{1,2}$, \; James S. Supancic III$^{1}$, \; Deva Ramanan$^{1}$ \\
$^{1}$Dept of Computer Science, University of California, Irvine, CA, USA\\
$^{2}$Universidad de Zaragoza, Zaragoza, Spain\\
{\tt\small grogez@unizar.es \;\; {grogez,jsupanci,dramanan}@ics.uci.edu} 
}

\maketitle

\begin{abstract}
  We tackle the problem of estimating the 3D pose of an individual's
  upper limbs (arms+hands) from a chest mounted
  depth-camera. Importantly, we consider pose estimation during
  everyday interactions with objects. Past work shows that strong
  pose+viewpoint priors and depth-based features are crucial for
  robust performance. In egocentric views, hands and arms are
  observable within a well defined volume in front of the
  camera. We call this volume an egocentric workspace. A notable property is that hand appearance correlates with workspace location. To exploit this correlation, 
  we classify arm+hand configurations in a global egocentric coordinate frame, rather than a local scanning window. This greatly simplify the architecture and improves performance.
We propose an efficient pipeline which 1) generates synthetic  workspace exemplars for training using a virtual chest-mounted camera whose intrinsic
  parameters match our physical camera, 2) computes
  perspective-aware depth features on this entire volume and 3)
  recognizes discrete arm+hand pose classes through a sparse multi-class
  SVM. Our method provides state-of-the-art hand
  pose recognition performance from egocentric RGB-D images in
  real-time.

\end{abstract}

\section{Introduction}
 Understanding hand poses and hand-object manipulations from a
 wearable camera has potential applications in assisted living~\cite{adl_cvpr12},
 augmented reality~\cite{FathiHR12} and life logging~\cite{LuG13}. As opposed to hand-pose
 recognition from third-person views, egocentric views may
 be more difficult due to additional occlusions 
 (from manipulated objects, or self-occlusions of fingers by the palm) 
 and the fact that hands interact with the environment and often leave the
 field-of-view.  The latter necessitates constant
   re-initialization, precluding the use of a large body of hand
   trackers which typically perform well given manual initialization. 
 \begin{figure}[h!]
\begin{tabular}{c}
  \centering 
  \hspace{-2mm}\includegraphics[width=0.40\textwidth]{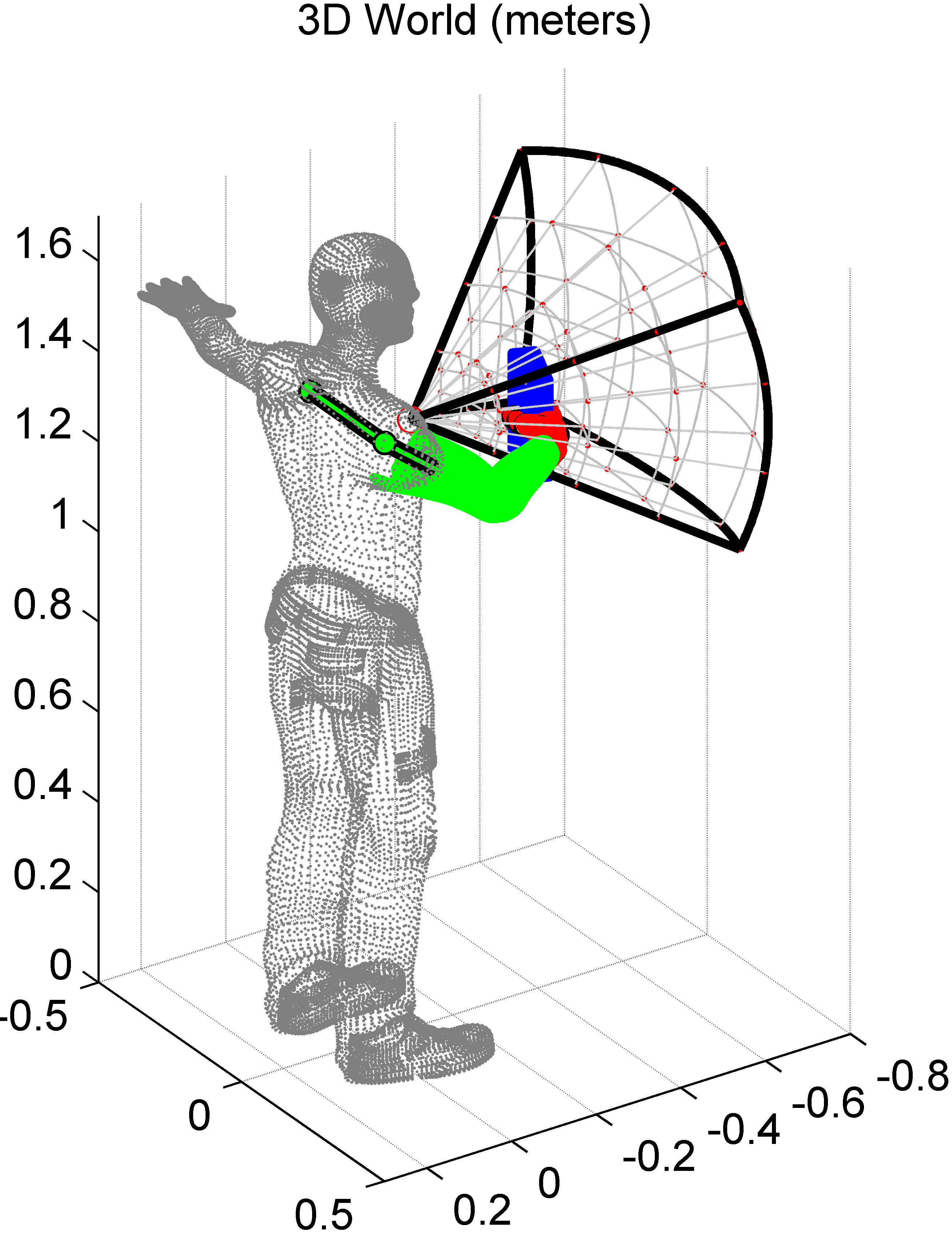}  
\end{tabular} 
\caption{{\bf Egocentric workspaces}.  We directly model the observable egocentric workspace in front of a human with a 3D volumetric descriptor, extracted from a 2.5D egocentric depth sensor. In this example, this volume is discretized into $4 \times 3 \times 4$ bins. This feature can be used to accurately predict shoulder, arm, hand poses, even when interacting with objects. We describe models learned from synthetic examples of observable egocentric workspaces obtained by place a virtual Intel Creative camera on the chest of an animated character.
  \label{fig:splash}}
\end{figure}


Previous work for egocentric hand analysis tends to rely on local 2D features, such as pixel-level skin classification \cite{LiICCV13,LiCVPR13} or gradient-based processing of depth maps with scanning-window templates~\cite{RogezKSMR2014}. Our approach follows in the tradition of ~\cite{RogezKSMR2014}, who argue that near-field depth measures obtained from a egocentric-depth sensor considerably simplifies hand analysis. Interestingly, egocentric-depth is not ``cheating'' in the sense that humans make use of stereoscopic depth cues for near-field manipulations~\cite{fielder1996does}. We extend this observation by building an explicit 3D map of the observable near-field workspace.

\begin{figure}
  \includegraphics[width=\columnwidth]{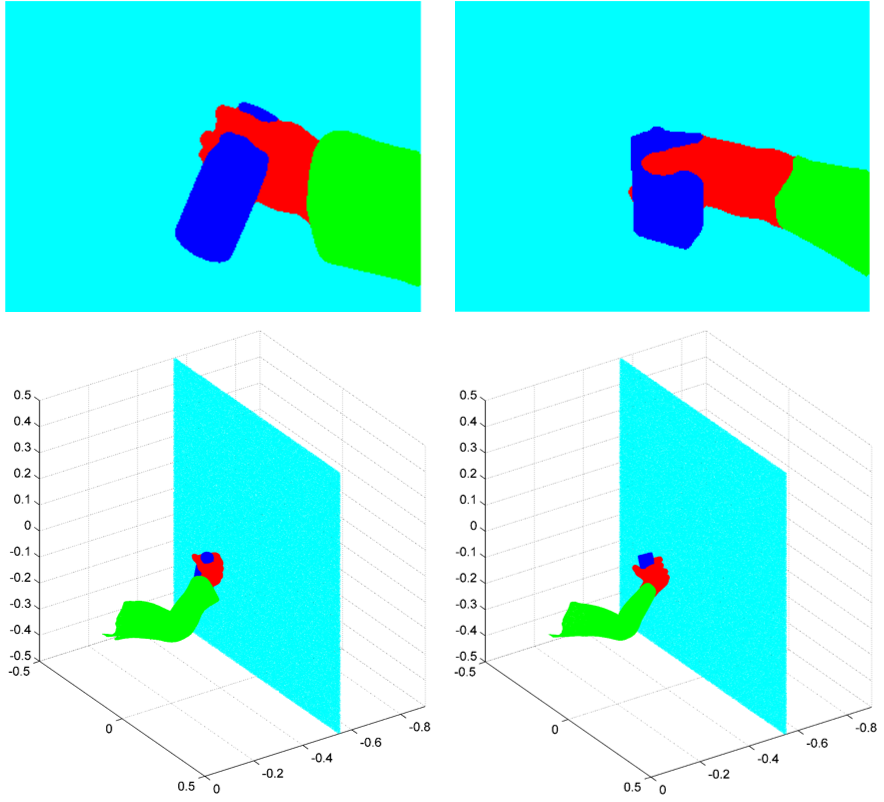}   
\caption{{\bf Problem statement.} The dimensionality of the problem is
  ${\color{green} N_{arm} } \times  {\color{red} N_{hand}}  \times
  {\color{blue} N_{object}} \times {\color{cyan} N_{background}}$. In
  this work, we will randomly sample ${\color{green} N_{arm} }$,
  consider a fixed set of hand-object configurations (${\color{red}
    N_{hand}}  \times {\color{blue} N_{object}} = 100$) and a fixed
  set of ${\color{cyan} N_{background}}$ background images captured
  with an Intel Creative depth camera. 
  To deal with the background, we will cluster the dataset and learn
  discriminative multi-classifiers, robust to background.  
  For each hand-object model, we randomly perturbed shoulder, arm and hand joint angles as physically possible to create a  new dense cloud of 3D points for arm+hand+object. We show 2 examples of hand-object models, a bottle (left) and a juice box (right) rendered in front of a flat wall.
  \label{fig:pose_examples}
}
\end{figure}

{\bf Our contributions:} In this work, we describe a new computational architecture that makes use of {\bf global} egocentric views, {\bf volumetric} representations, and {\bf contextual} models of interacting objects and human-bodies. Rather than detecting hands with a local (translation-invariant) scanning-window classifier, we process the entire global egocentric view (or {\em work-space}) in front of the observer (Fig.~\ref{fig:splash}).  Hand appearance is not translation-invariant due to perspective effects and kinematic constraints with the arm. To capture such effects, we build a library of synthetic  3D egocentric workspaces generated using real capture conditions (see examples in Fig.~\ref{fig:pose_examples}). We animate a 3D human character model inside virtual scenes with objects, and render such animations with a chest-mounted camera whose intrinsics match our physical camera . We simultaneously recognize arm and hand poses while interacting with objects by classifying the whole 3D volume using a multi-class Support Vector Machine (SVM) classifier. Recognition is simple and fast enough to be implemented in 4 lines of code.

\subsection{Related work} 
\label{sec:litr}
 
{\bf Hand-object pose estimation:}
While there is a large body of work on
hand-tracking~\cite{KurataKKJE02,KolschT05,Kolsch10,Athitsos03estimating3d,Oikonomidis2012,StengerPAMI06,wang2009rth},
we focus on hand pose estimation during object manipulations. Object
interactions both complicate analysis due to additional occlusions,
but also provide additional contextual constraints (hands cannot
penetrate object geometry, for example). \cite{hamer_tracking_2009}
describe articulated tracker with soft anti-penetration constraints,
increasing robustness to occlusion. Hamer \emph{et al.} describe
contextual priors for hands in relation to
objects~\cite{hamer_object-dependent_2010}, and demonstrate their
effectiveness for increasing tracking accuracy. Objects are
  easier to animate than hands because they have fewer joint
  parameters. With this intuition, object motion can be used as an input signal for estimating hand motions \cite{hamer_data-driven_2011}. \cite{RomeroKEK13} use a large synthetic dataset of hands manipulating objects, similar to us. We differ in our focus on single-image and egocentric analysis.



{\bf Egocentric Vision:}
Previous egocentric studies have focused on activities of daily
living~\cite{adl_cvpr12,fathiunderstanding}. Long-scale temporal
structure was used to handle complex hand object interactions,
exploiting the fact that objects look different when they are
manipulated (active) versus not manipulated (passive)
\cite{adl_cvpr12}. Much previous work on egocentric hand recognition
make exclusive use of RGB cues \cite{LiCVPR13,limodel}, while we focus
on volumetric depth cues. Notable exceptions include
\cite{DamenGMC12}, who employ egocentric RGB-D sensors for personal
workspace monitoring in industrial environments and
~\cite{MannHJLRCD11}, who employ such sensors to assist blind users in
navigation.

{\bf Depth features:} Previous work has shown the efficacy of depth
cues~\cite{shotton_efficient_2013, XuChe_iccv13}. 
We compute volumetric depth features from point clouds.
Previous work has examined point-cloud processing of
depth-images~\cite{YeWYRP11,shotton2013real,XuChe_iccv13}. A common
technique estimates local surface orientations and normals
\cite{YeWYRP11,XuChe_iccv13}, but this may be sensitive to noise since
it requires derivative computations. We employ simpler volumetric
features, similar to \cite{song2014sliding} except that we use a spherical coordinate frame that does not slide along a scanning window (because we want to measure depth in an egocentric coordinate frame).

 

\noindent 
{\bf Non-parametric recognition:} Our work is inspired by non-parametric techniques that make use of synthetic training data \cite{RomeroKEK13,shakhnarovich2003fast,hamer_tracking_2009,transduc:forest,tzionas_comparison_2013}. \cite{shakhnarovich2003fast} make use of pose-sensitive hashing techniques for efficient matching of synthetic RGB images rendered with Poser. We generate synthetic depth images, mimicking capture conditions of our actual camera. 

 \begin{figure}
\begin{tabular}{cc}
  \centering 
  \hspace{-2mm}\includegraphics[trim=0.25cm 0.5cm 0.5cm 0.05cm, clip=true,width=0.23\textwidth]{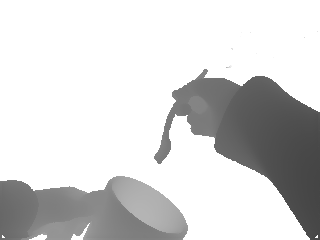} &
  \hspace{-3mm}\includegraphics[trim=0.25cm 0.5cm 0.5cm 0.05cm,clip=true,width=0.23\textwidth]{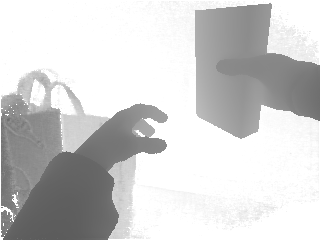} \\
  (a)&(b)\\
  \hspace{-3mm}\includegraphics[trim=0.25cm 0.5cm 0.5cm 0.05cm,clip=true,width=0.23\textwidth]{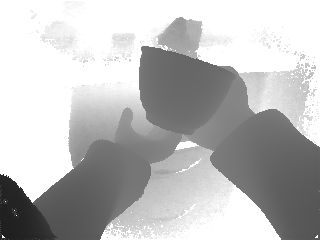} &
  \hspace{-2mm}\includegraphics[trim=0.25cm 0.5cm 0.5cm 0.05cm,clip=true,width=0.23\textwidth]{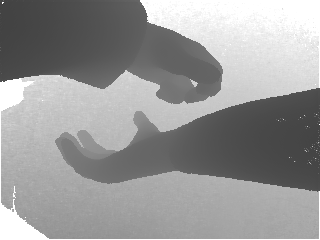} \\
  (c)&(d)
\end{tabular} 
\caption{{\bf Examples of synthetic training images}. We show examples of training depth images produced with our rendering procedure. Surprisingly,  realistic multi-arm configurations are generated as depicted in these 4 examples (a-d) where two hands manipulate household objects with a realistic random background.  
  \label{fig:trainingdata}
}
\end{figure}

\section{Training data}

 The dataset employed in this paper is made of realistic synthetic 3D exemplars which are generated simulating real capture conditions:   synthetic 3D hand-object data, rendered with a 3D computer graphics program, are combined with real 3D background scenario and  rendered using the test camera  projection matrix.
 
 {\bf Poser models.} Our synthetic database includes more than 200
 different grasping hand postures with and without objects. We also
 varied the objects being interacted with, as well as the clothing of
 the character, i.e., with and without sleeves. Overall we used 49
 objects, including kitchen utensils, personal bathroom items,
 office/classroom objects, fruits, etc. Additionally we used 6
   models of empty hands: waive, fist, thumbs-up, point, etc . Note that
   some objects can be handled with different postures. For instance,
    {when we open a bottle we do not use the same posture (to grasp its
     cap and neck) as we do to idly grasp its body .   
   We added several such  variant models to our database, i.e., different hand postures  manipulating  the
   same object.  
 
 {\bf Kinematic model.} Let $\theta$ be a vector of arm joint angles, and let $\phi$ be a vector of grasp-specific hand joint angles, obtained from the above set of Poser models. We use a standard forward kinematic chain to convert the location of finger joints ${\bf u}$ (in a local coordinate system) to image coordinates:
\begin{align}
&  {\bf p} = C \prod_i T(\theta_i) \prod_j T(\phi_j) {\bf u}, \quad \text{where} \quad T,C \in \mathbb{R}^{4 \times 4}, \nonumber \\
&{\bf u} = \begin{bmatrix} u_x & u_y & u_z & 1 \end{bmatrix}^T, \quad (x,y) = (f \frac{p_x}{p_z}, f\frac{p_y}{p_z}), \label{eq:kinematics}
\end{align}
 \noindent where $T$ specifies rigid-body transformations (rotation and translation) along the kinematic chain and $C$ specifies the extrinsic camera parameters. Here {\bf p} represents the 3D position of point {\bf u} in the camera coordinate system. To generate the corresponding image point, we assume camera intrinsics are given by identity scale factors and a focal length $f$ (though it is straightforward to use more complex intrinsic parameterizations). We found it important to use the $f$ corresponding to our physical camera, as it is crucial to correctly model perspective effects for our near-field workspaces. 

{\bf Viewpoint-dependent translations:} 
  We wish to enrich the core set of posed hands with additional translations and viewpoints. The parametrization of visible arm+hand configurations is non-trivial. To do so, we take a simple {\em rejection sampling} approach. We fix $\phi$ parameters to respect the hand grasps from Poser, 
and add small Gaussian perturbations to arm joint angles $$\theta_i' = \theta_i + \epsilon \quad \text{where} \quad \epsilon \sim N(0,\sigma^2).$$ Importantly, this generates hand joints {\bf p} at different translations and viewpoints, correctly modeling the dependencies between both. For each perturbed pose, we render hand joints using \eqref{eq:kinematics} and keep poses where 90\% of them are visible (e.g., their $(u,v)$ coordinate lies within the image boundaries). 


{\bf Depth maps.} 
Associated with each rendered set of keypoints, we would also like a depth map. To construct a depth map, we represent each rigid limb with a dense cloud of 3D vertices $\{{\bf u}_i\}$. We produce this cloud by (over) sampling the 3D meshes defining each rigid-body shape. We render this dense cloud using forward kinematics \eqref{eq:kinematics}, producing a set of points $\{{\bf p}_{i}\}=\{(p_{x,i},p_{y,i},p_{z,i})\}$.
We define a 2D depth map $z[u,v]$ by ray-tracing. Specifically, we cast a ray from the origin, in the direction of each image (or depth sensor) pixel location $(u,v)$ and find the closest point:
\begin{align}
 z[u,v] = \min_{k \in \text{Ray}(u,v)}||{\bf p}_k|| \label{eq:depth}
\end{align}
\noindent where $\text{Ray}(u,v)$ denotes the set of points on (or near) the ray passing through pixel $(u,v)$. We found the above approach simpler to implement than hidden surface removal, so long as we projected a sufficiently dense cloud of 3D points.

{\bf Multiple hands:} Some object interactions require multiple hands interacting with a single object. Additionally, many views contain the second hand in the ``background''. For example, two hands are visible in roughly 25\% of the frames in our benchmark videos.  We would like our training dataset to have similar statistics. Our existing Poser library contains mostly single-hand grasps. To generate additional multi-arm egocentric views, we randomly pair 25\% of the arm poses with a mirrored copy of another randomly-chosen pose. We then add noise to the arm joint angles, as described above. Such a procedure may generate unnatural or self-intersecting poses. To remove such cases, we separately generate depth maps for the left and right arms, and only keep pairings that produce compatible depth maps:
\begin{align}
  | z_{left}[u,v] - z_{right}[u,v] | > \delta \quad \forall u,v
\end{align}
We find this simple procedure produces surprisingly realistic multi-arm configurations (Fig.~\ref{fig:trainingdata}). Finally we add background clutter from depth maps of real egocentric scenes (not from our benchmark data). We used the above approach to produce a dataset of 500,000 multi-hand(+arm+objects) configurations and associated depth-maps. 
  \begin{figure*}[t!]
  \centering
 \includegraphics[width=\textwidth]{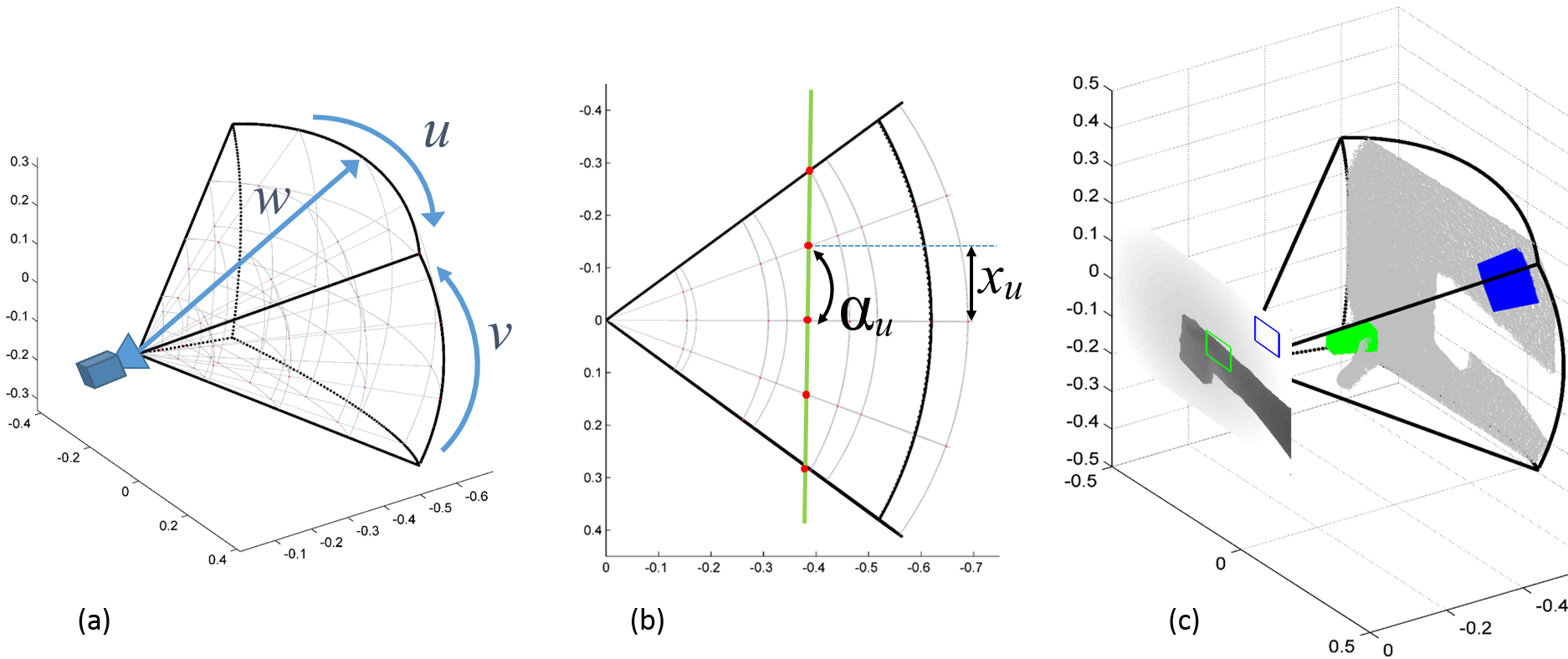} 
\caption{{\bf Volume quantization.}  We  quantize those points that fall within the egocentric workspace in front of the camera (observable volume within $z_{max} = 70 \text{cm}$) into a binary spherical voxel grid of $N_u \times N_v \times N_w$ voxels (a). We vary the azimuth angle $\alpha$ to generate equal-size projections on the image plane (b). Spherical bins ensure that voxels at different distances project to same image area (c). This allows for efficient feature computation and occlusion handling, since occluded voxels along the same line-of-sight can be identified by iterating over $w$.
\label{fig:Vol}  
\label{fig:Frustum}
}
\end{figure*}

\section{Formulation}

\subsection{Perspective-aware depth features}

Objects close to the lens appear large relative to more distant objects and cover greater areas of the depth map. Much previous work has proposed to remove  the effect of the perspective projection  by computing depth feature in real-world orthographic space, e.g. by quantizing 3D points clouds, for instance to train translation-invariant detectors. We posit that perspective distortion is useful in egocentric settings and should be exploited: objects of interest (hands, arms, and manipulated things) tend to lie near the body and exhibit perspective effects. To encode such phenomena, we construct a spherical bin histogram by gridding up the egocentric workspace volume by varying azimuth and elevation angles (See Fig.~\ref{fig:Frustum}).  We demonstrate that this feature performs better than orthographic counterparts, and is also faster to compute.



 

 \begin{figure*} 
    \hspace{-4mm}\includegraphics[width=\textwidth]{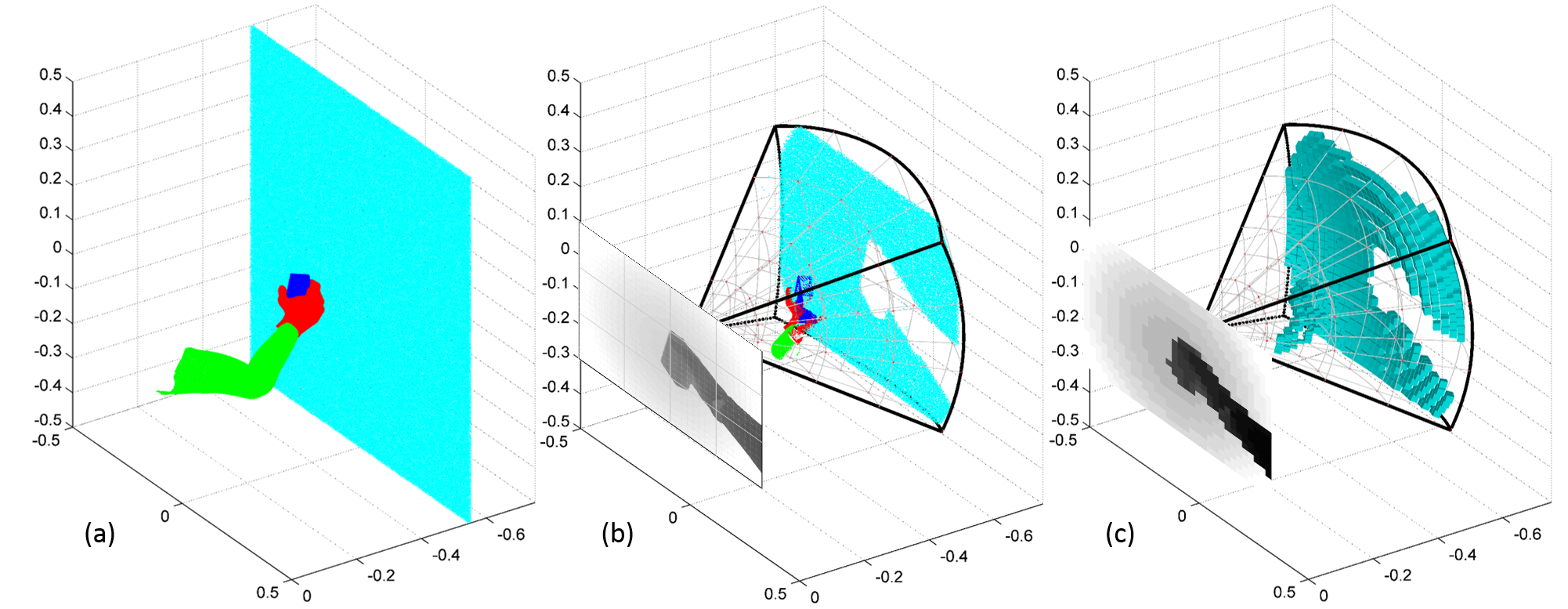}  
\caption{ {\bf Binarized volumetric feature.} We synthesize training examples by randomly perturbing shoulder, arm and hand joint angles in a physically possible manner (a). For each example, a synthetic depth map is created by projecting the visible set of dense 3D point clouds using a real-world camera projection matrix (b). The resulting 2D depth map is then quantized with a regular grid in x-y directions and binned in the viewing direction to compute our new binarized volumetric feature (c). In this example, we use a $32\times24\times35$ grid. Note that for clarity we only show the sparse version of our 3D binary feature. We also show the quantized depth map $z[u,v]$ as a gray scale image (c). 
  \label{fig:feature}
}
\end{figure*} 
{\bf Binarized volumetric features:} Much past work processes depth maps as 2D rasterized sensor data. Though convenient for applying efficient image processing routines such as gradient computations (e.g., \cite{tang2013histogram}), rasterization may not fully capture the 3D nature of the data. Alternatively, one can convert depth maps to a full 3D point cloud~\cite{LaiBF14}, but the result is orderless making operations such as correspondence-estimation difficult. We propose encoding depth data in a 3D volumetric representation, similar to \cite{song2014sliding}.  To do so, we can back-project the depth map from \eqref{eq:depth} into a cloud of visible 3D points $\{{\bf p}_{k}\}$, visualized in Fig.~\ref{fig:feature}-(b). They are a subset of the original cloud of 3D points $\{{\bf p}_{i}\}$ in Fig.~\ref{fig:feature}-(a).  We now bin those visible points that fall within the egocentric workspace in front of the camera (observable volume within $z_{max} = 70 \text{cm}$) into a binary voxel grid of $N_u \times N_v \times N_w$ voxels:
\begin{align}
  b[u,v,w] = \left\{
  \begin{array}{c l l}      
    1 & \text{if} & \exists k \text{  s.t.} \quad
    {\bf p}_{k} \in F(u,v,w)\\
    0 & & \text{otherwise}
  \end{array}\right.  
\end{align}
\noindent where $F(u,v,w)$ denotes the set of points within a voxel centered at coordinate $(u,v,w)$. 

{\bf Spherical voxels:} Past work tends to use rectilinear voxels~\cite{song2014sliding,LaiBF14}. Instead, we use a spherical binning structure, centering the sphere at the camera origin (~Fig.~\ref{fig:Vol}). At first glance, this might seem strange because voxels now vary in size -- those further away from the camera are larger. The main advantage of a ``perspective-aware'' binning scheme is that all voxels now project to the same image area in pixels (Fig.~\ref{fig:Vol}-(c)}. This in turn makes feature computation extremely efficient, as we will show.

{\bf Efficient quantization:} Let us choose spherical bins $F(u,v,w)$ such that they project to a single pixel $(u,v)$ in the depth map. This allows one to compute the binary voxel grid $b[u,v,w]$ by simply ``reading off'' the depth value for each $z(u,v)$ coordinates, quantizing it to $z'$, and assigning 1 to the corresponding voxel: 
\begin{align}
  b[u,v,w] = \left\{
  \begin{array}{l l l}      
    1 & \text{if} &\quad w  = z'[u,v]\\
    0 & & \text{otherwise}
  \end{array}\right.  \label{eq:voxel}
\end{align}


This results in a sparse volumetric voxel features visualized in Fig.~\ref{fig:feature}-(c). Once a depth measurement is observed at position $b[u',v',w'] = 1$, all voxels behind it are occluded for $w\geq w'$. This arises from the fact that single camera depth measurements are, in fact, 2.5D. By convention, we define occluded voxels to be ``1''.

In practice, we consider a coarse discretization of the volume to make the problem more tractable. The depth map $z[x,y]$ is resized to $N_u \times N_v$ (smaller than depth map size) and quantized in $z$-direction. To minimize the effect of noise when counting the points which fall in the different voxels, we quantize the depth measurements by applying a median filter on the pixel values within each image region:
\begin{align}
\begin{array}{c}
\forall u,v\in [1, N_u]\times[1, N_v],  \\ z'[u,v]=\frac{N_w}{z_{max}}\text{median}(z[x,y]:(x,y)\in P(u,v)),
\end{array}
\end{align}
where $P(u,v)$ is the set of pixel coordinates in the original depth map corresponding to pixel coordinate $(u,v)$ coordinates in the resized depth map.




\subsection{Global pose classification}
\label{sec:globalpose}

We quantize the set of poses from our synthetic database into $K$ coarse classes for each limb, and train a $K$-way pose-classifier for pose-estimation. The classifier is linear and makes use of our sparse volumetric features, making it quite simple and efficient to implement.

\begin{figure*} 
  \hspace{-4mm}\includegraphics[width=\textwidth]{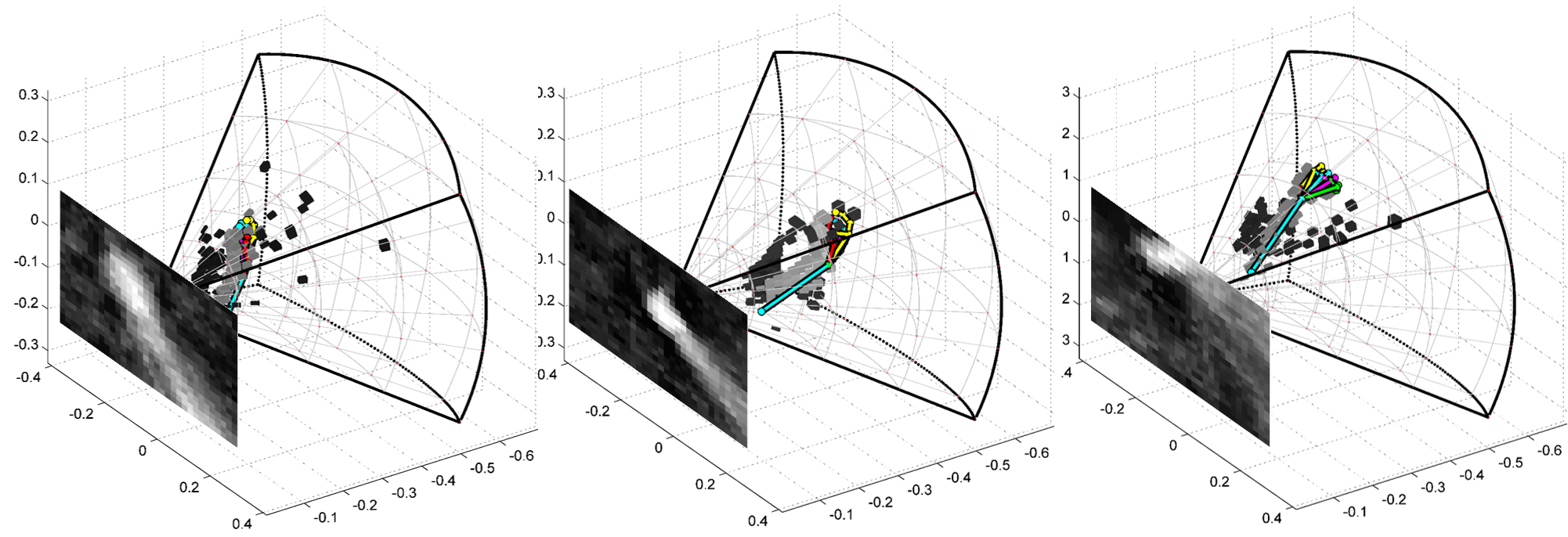}    
\caption{{\bf Pose classifiers.} We visualize the linear weight tensor $\beta_k[u,v,w]$ learnt by the SVM for a $32\times24\times35$ grid of binary features for 3 different pose clusters. We plot a 2D $(u,v)$ visualization obtained by computing the max along $w$. We also visualize the corresponding average 3D pose in the egocentric volume together with the top 500 positive (light gray) and negative weights (dark gray) within $\beta_k[u,v,w]$. 
  \label{fig:PoseClusters}
}
\end{figure*}

{\bf Pose space quantization: } For each training exemplar, we generate the set of 3D keypoints: 17 joints (elbow + wrist + 15 finger joints) and the 5 finger tips. Since we want to recognize coarse limb (arm+hand) configurations, we cluster the resulting training set by applying K-means to the elbow+wrist+knuckle 3D joints. We usually represent each of the K resulting clusters using the average 3D/2D keypoint locations of both arm+hand (See examples in Fig.~\ref{fig:PoseClusters}). Note that K can be chosen as a compromise between accuracy and speed.

{\bf Global classification:} We use a linear SVM for a multi-class classification of upper-limb poses. However, instead of classifying local scanning-windows, we classify global depth maps quantized into our binarized depth feature $ b[u,v,w]$ from \eqref{eq:voxel}. Global depth maps allow the classifier to exploit contextual interactions between multiple hands, arms and objects. In particular, we find that modeling arms is particularly helpful for detecting hands. For each class $k \in \{1, 2, ... K  \}$, we train a one-vs-all SVM classifier obtaining weight vector which can be re-arranged into a $N_u \times N_v \times N_w$ tensor $\beta_k[u,v,w]$. The score for class k is then obtained by a simple dot product of this weight and our binarized feature $b[u,v,w]$:
\begin{align}
  \text{score}[k]= \sum_{u} \sum_{v}  \sum_{w}  \beta_k[u,v,w] \cdot b[u,v,w].  \label{eq:svm}
\end{align}
In Fig.~\ref{fig:PoseClusters}, we show the weight tensor $\beta_k[u,v,w]$ for 3 different pose clusters. 
 
\subsection{Joint feature extraction and classification} 

To increase run-time efficiency, we exploit the sparsity of our binarized volumetric feature and jointly implement feature extraction and SVM scoring. Since our binarized depth features do not require any normalization and the classification score is a simple dot product, we can readily extract the feature and update the score on the fly. 

Because all voxels behind the first measurement are backfilled, the SVM score for each class $k$ from \eqref{eq:svm} can be written as:
\begin{align}
  \text{score}[k]= \sum_{u} \sum_{v} \beta_k'[u,v,z'[u,v]],  \label{eq:eff}
\end{align}
\noindent where $z'[u,v]$ is the quantized depth map and tensor $\beta_k'[u,v,w]$ is the cumulative sum of the weight tensor along dimension $w$:
\begin{align}
  \beta'_k[u,v,w] = \sum_{d >= w} \beta_k[u,v,d]
\end{align}
Note that the above cumulative-sum tensors can be precomputed. This makes test-time classification quite efficient \eqref{eq:eff}. Feature extraction and SVM classification can be computed jointly following the algorithm presented in  Alg.~\ref{alg:classif}. We invite the reader to view our code in supplementary material. 
\begin{algorithm}[htb]
\SetAlFnt{\tiny\sf}
\SetKwInOut{Input}{input} \SetKwInOut{Output}{output}

\Input{Quantized depth map $z'[u,v]$.\\ Cumsum'ed weights $\{\beta_k'[u,v,w]\}$.

\Output{$\text{score}[k]$}
\BlankLine  
\For {$  u \in \{0,1, ...  N_u\}$}
{\For {$  v \in \{0,1, ...  N_v\}$} 
{\For {$  k \in \{0,1, ...  K\}$}
{$\text{score}[k] += \beta_k'[u,v,z'[u,v]] $}
} 
}
} 
\caption{Joint feature extraction and classification. We jointly extract binarized depth features and evaluate linear classifiers for all quantized poses $k$. We precompute a ``cumsum'' $\beta_k'$ of our SVM weights. At each location $(u,v)$, we add all the SVM weights corresponding to the voxels behind $z[u,v]$, i.e. such that $w\geq z[u,v]$.}
\label{alg:classif} 
\end{algorithm}
 
\section{Experiments}

For evaluation, we use the recently released UCI Egocentric dataset~\cite{RogezKSMR2014} and score hand pose detection as a proxy for limb pose recognition (following the benchmark criteria used in ~\cite{RogezKSMR2014}) . The dataset consists of 4 video sequences (around 1000 frames each) of everyday egocentric scenes with hand annotations every 10 frames. Our unoptimized matlab implementation runs at 15 frames per second. 
 \begin{figure}[h!]
\begin{tabular}{cc}
  \centering 
  \footnotesize  Feature comparison  & \footnotesize Feature Resolution \\
  \hspace{-2mm}\includegraphics[width=0.5\columnwidth]{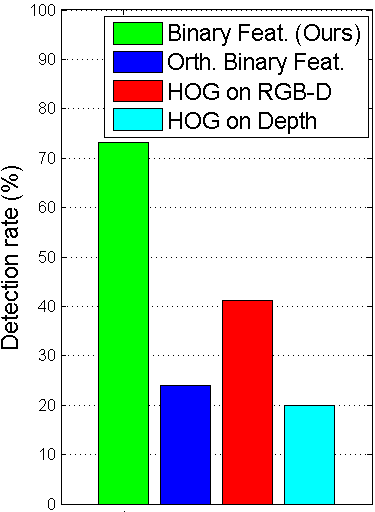}  & 
  \hspace{-4mm}\includegraphics[width=0.5\columnwidth]{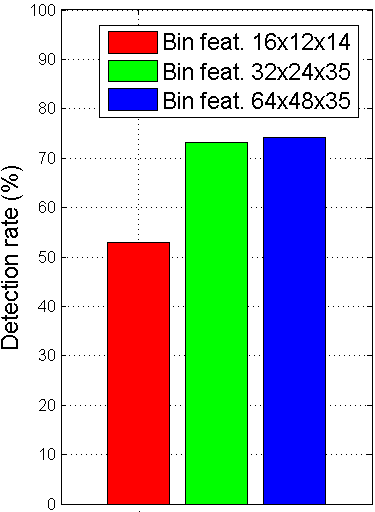} \\
  (a)  &(b)
\end{tabular} 
\caption{{\bf Feature evaluation}. We compare different type of features, volumetric features, HOG on RGB-D, HOG on Depth for $K=750$ classes (a). For our perspective binary features and the orthographic binary features, we consider regular grids of dimensions $32\times24\times35$. For HOG on depth and HOG on RGB-D, we  respectively use $30\times40$  and $16\times24$ cells  with 16 orientation bins. Our perspective binary features clearly outperforms other types of features. We also show results varying the resolution of our proposed feature in (b), again $K=750$. We can observe how $32\times24\times35$ is a good trade-off between feature dimensionality and performance, which validates our choice. Doubling the resolution in $u,v$ marginally improves accuracy.
  \label{fig:EvalFeatures}}
\end{figure}

 
{\bf Feature evaluation:}
We first compare hand detection accuracy for different K-way SVM classifiers trained on HOG on depth (as in \cite{RogezKSMR2014} ) and HOG on RGB-D, thus exploiting the stereo-views provided by RGB and depth sensors. To show the benefit of preserving the perspective when encoding depth features, we also experimented with an orthographic version of our binarized volumetric feature (similar to past work ~\cite{song2014sliding,LaiBF14}). In that case, we quantize those points that fall within a 64x48x70 $cm^3$ egocentric workspace in front of the camera into a binary voxel grid:
\begin{align}
  b_{\perp}[u,v,w] = \left\{
  \begin{array}{c l l}      
    1 & \text{if} & \exists i \text{  s.t.} \quad (x_i,y_i,z_i) \in N(u,v,w)\\
    0 & & \text{otherwise}
  \end{array}\right.  
\end{align}
\noindent where $N(u,v,w)$ specifies a $2 \times 2 \times 2 cm$ cube centered at voxel $(u,v,w)$. Note that this feature is considerable more involved to calculate, since it requires an explicit backprojection and explicit geometric computations for binning. It is also not clear how to identify occluded voxels because they are not arranged along line-of-sight rays. 

The results obtained with $K=750$ pose classes are reported in Fig.~\ref{fig:EvalFeatures}-(a). Our perspective binary features clearly outperforms other types of features. We reach $72\%$ detection accuracy while state of the art algorithm~\cite{RogezKSMR2014} reports $60\%$ accuracy. Our volumetric feature has empirically strong performance in egocentric settings.
One reason is that it is robust to small intra-cluster misalignment
and deformations because all voxels behind the first measurement are backfilled.
Second, it is sensitive to variations in apparent size induced by perspective effects (because voxels have consistent perspective projections). In Fig.~\ref{fig:EvalFeatures}-(b), we also show results varying the resolution of the grid. Our choice of $32\times24\times35$ is a good trade-off between feature dimensionality and performance.
 \begin{figure}[htb!]
\begin{tabular}{ccc}
  \centering 
   \footnotesize  Detection varying $K$& \footnotesize Detection varying size of training \\
  \hspace{-4mm}\includegraphics[width=0.5\columnwidth]{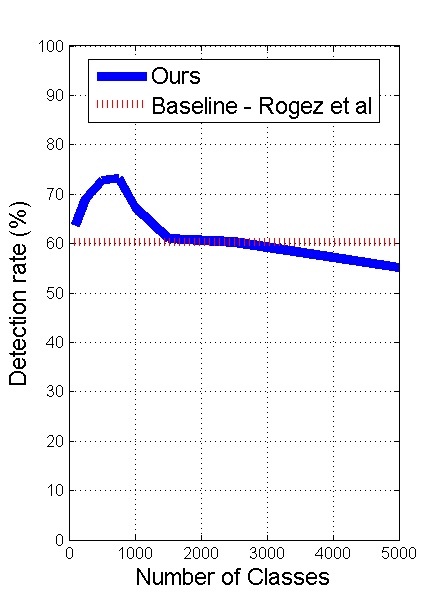} & 
  \hspace{-3mm}\includegraphics[width=0.5\columnwidth]{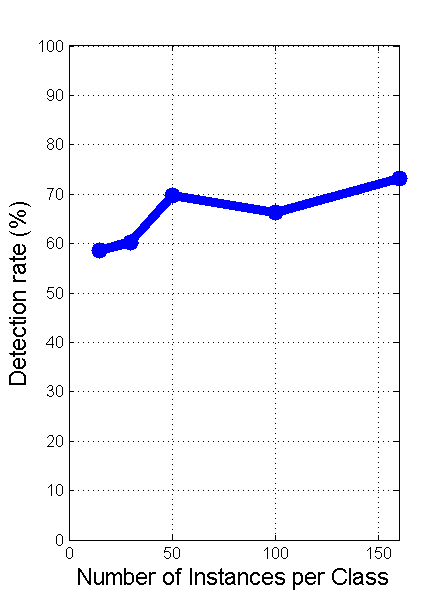} \\
  (a)  &(b)
\end{tabular} 
\caption{{\bf Clustering and size of training set.} We compare hand pose detection against state of the art method \cite{RogezKSMR2014} in (a). The results are given varying the number of discretized pose classes for a total of 120,000 training exemplars. We see that we reach a local maxima for $K=750$. In (b), we show how increasing the number of positive training exemplars used to train each 1-vs-all SVM classifier slowly increases accuracy. 
  \label{fig:Eval}}
\end{figure}

{\bf Training data and clustering:} 
We evaluated the performance of our algorithm when varying the discretization of a set of 120,000 training images, i.e. varying the number of pose classes. We can observe in Fig.~\ref{fig:Eval}-(a) that we reach a local maxima for $K=750$. This suggests that for $K\geq 750$ there is not enough training data to train robust SVM classifiers and our model over-fits. We trained several $K$-way classifiers varying the number of training instances for each class. Increasing the number of positive training exemplars used to train each 1-vs-all SVM classifier slowly increases accuracy as shown in Fig.~\ref{fig:Eval}-(b). These results suggest that a massive training data set and a finer quantization of the pose space ($K\geq 750$) should outperform our existing model.
 
\begin{figure*}[htb!]
  \centering
\includegraphics[width=\textwidth]{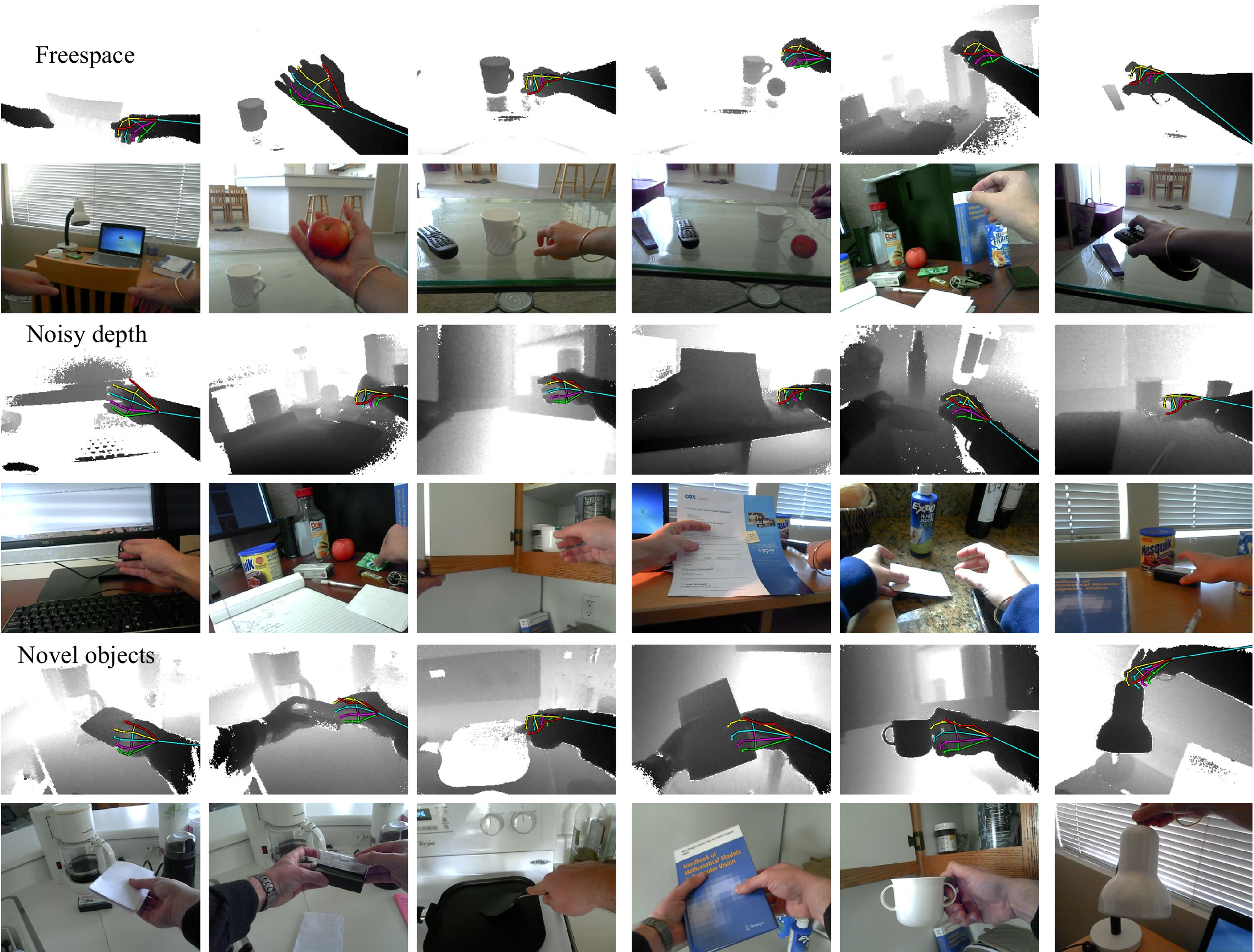}
\caption{{\bf Good detections}.  We show frames where arm and  hand are correctly detected. First, we present some easy cases of hands in free-space ({\bf top row} ). Noisy depth data and cluttered background cases ({\bf middle row}) showcases the robustness of our system while novel objects ({\bf bottom row:} envelope, staple box, pan, double-handed cup and lamp) require generalization to unseen objects at train-time.  
  \label{fig:easycases}
}
\end{figure*}

\begin{figure*}[htb!]
  \centering
  \includegraphics[width=\textwidth]{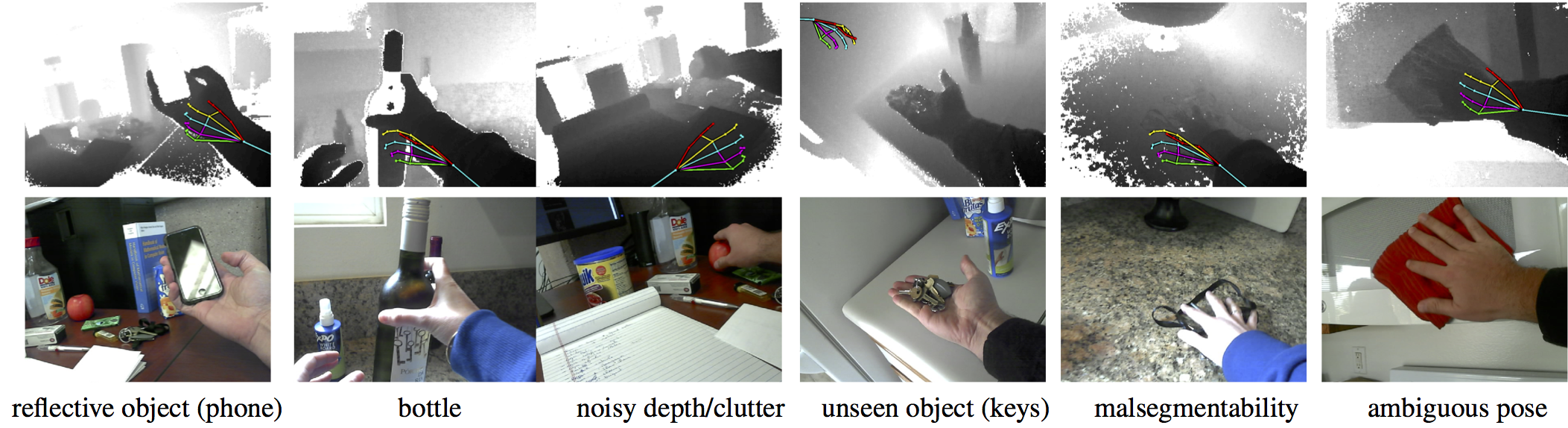}
\caption{{\bf Hard cases}. We show frames where the pose is not correctly recognized ( sometimes not even detected) by our system. These hard cases include excessively-noisy depth data, hands manipulating reflective material (phone  or bottle of wine), malsegmentability cases of hands touching background.
  \label{fig:hardcases}
}
\end{figure*}

{\bf Qualitative results:} We illustrate successes in difficult scenarios in Fig.~\ref{fig:easycases} and analyze common failure modes in Fig.~\ref{fig:hardcases}. Please see the figures for additional discussion. We also invite the reader to view our supplementary videos for additional results. 

\section{Conclusions}

We have proposed a new approach to the problem of egocentric 3D hand pose recognition during  interactions with objects. Instead of classifying local depth image regions through a typical translation-invariant scanning window, we have shown that classifying the global arm+hand+object configurations within the ``whole'' egocentric workspace in front of the camera allows for fast and accurate results.  We train our model by synthesizing workspace exemplars consisting of hands, arms, objects and backgrounds. Our model explicitly reasons about perspective occlusions while being both conceptually and practically simple to implement (4 lines of code). We produce state-of-the-art real-time results for egocentric pose estimation.


{\small
\bibliographystyle{ieee}
\bibliography{deva_bib}
}

\end{document}